\documentclass[runningheads]{llncs}
\usepackage{graphicx}
\usepackage{amsmath,amssymb} 
\usepackage{color}\usepackage[width=122mm,left=12mm,paperwidth=146mm,height=193mm,top=12mm,paperheight=217mm]{geometry}

\usepackage{setspace}
\usepackage{authblk}

\usepackage{url}

\begin{document}
\pagestyle{headings}
\mainmatter

\title{Learning Image Matching by\\Simply Watching Video\thanks{Contact: gucan.long@nudt.edu.cn, laurent.kneip@anu.edu.au}} 

\authorrunning{Gucan Long, Laurent Kneip, Jose M. Alvarez, Hongdong Li}

\titlerunning{Learning Image Matching by Simply Watching Video}

\author{
Gucan Long\inst{1, 2} 
\and Laurent Kneip\inst{2, 4}
\and Jose M. Alvarez\inst{2, 3}
\and Hongdong Li\inst{2, 3, 4}
}

\institute{
National University of Defense Technology, P.R. China,
\and
Australian National University, Australia
\and
NICTA, Australia
\and
ARC Centre of Excellence for Robotic Vision, Australia}

\maketitle

\begin{abstract}

This work presents an unsupervised learning based approach to the ubiquitous computer vision problem of image matching. We start from the insight that the problem of frame-interpolation implicitly solves for inter-frame correspondences. This permits the application of analysis-by-synthesis: we firstly train and apply a Convolutional Neural Network for frame-interpolation, then obtain correspondences by inverting the learned CNN. The key benefit behind this strategy is that the CNN for frame-interpolation can be trained in an unsupervised manner by exploiting the temporal coherency that is naturally contained in real-world video sequences. The present model therefore learns image matching by simply ``watching videos". Besides a promise to be more generally applicable, the presented approach achieves surprising performance comparable to traditional empirically designed methods.

\end{abstract}

\section{Introduction}

\begin{figure}[t]
		\centering
		\includegraphics[width=0.7\columnwidth]{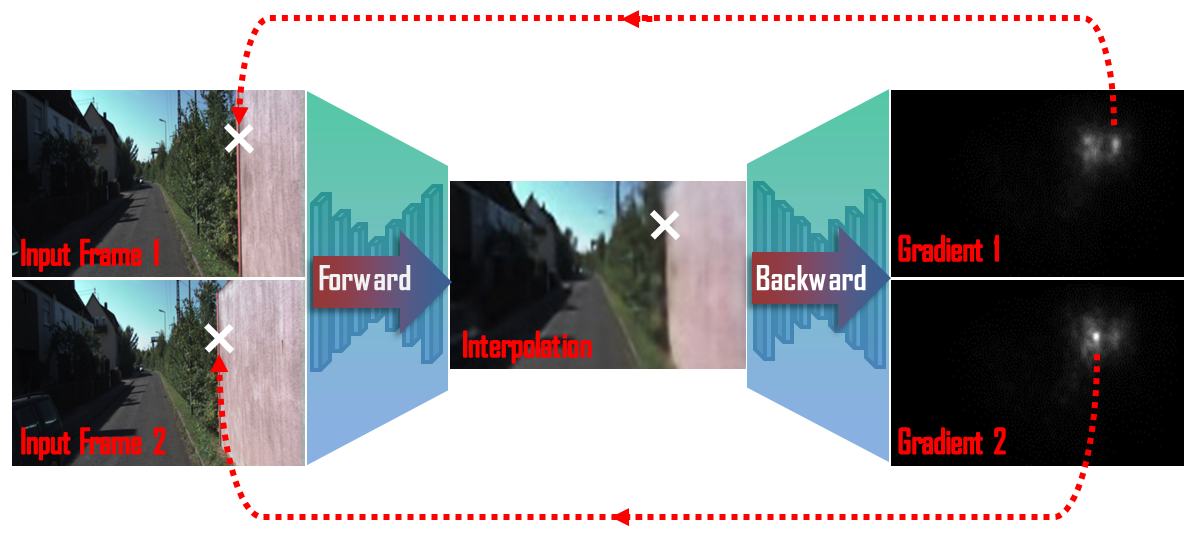}
		\caption{We train a deep convolutional network for frame interpolation, which can be done without manual supervision by exploiting the temporal coherency that is naturally contained in real-world video sequences. The learned CNN is then used to compute a sensitivity map for each output pixel. This sensitivity map, i.e. the gradients w.r.t. the input, indicates how much each input pixel influences a particular output pixel. The two input pixels (one per input frame) that have the maximum influence are considered as a match. Though indirect, the present model learn how to perform dense correspondence matching by simply watching video.}
		\label{fig:introduction}
	\end{figure}
	
	

  
  For our human beings, vision, and how the brain uses visual information, are learned skills. Meanwhile, the ultimate goal of computer vision research is to teach machines to understand the visual world. But obviously we cannot do it all in the manner of hand over hand, i.e. via empirically man-devised models. It would be more ideal and practicable if we can teach them to learn vision by themselves. This work focuses on the fundamental problem of establishing 2D-2D correspondences across a pair of consecutive frames, and notably proves that a solution to this low-level vision problem could be achieved in an unsupervised way by relying only on natural video sequences. 
  
  Our key insight lies in the understanding that frame interpolation implicitly solves for dense correspondences between the input image pair. It is well known that dense matching can be regarded as a sub-problem of frame-interpolation, as the interpolation could be immediately generated by correspondence-based image warping once dense inter-frame matches are available. It then comes as no surprise that if we were able to train a deep neural network for frame interpolation, its application would implicitly also generate knowledge about dense image correspondences. Retrieving this knowledge is known as \textit{analysis by synthesis} \cite{yildirim15}, a paradigm in which learning is described as the acquisition of a measurement synthesising model, and inference of generating parameters as model inversion once correct synthesis is achieved. In our context, \textit{synthesis} simply refers to frame interpolation. We then, for the \textit{analysis} part, show that the correspondences can be recovered from the network through gradient back-propagation, which produces sensitivity maps for each interpolated pixel. The procedure is summarised in Figure \ref{fig:introduction}, explaining how the reciprocal mapping between frame-interpolation and dense correspondences is encoded in the forward and backward propagation through one and the same network architecture. We call our approach MIND, which stands for Matching by INverting \footnote{The term of \textit{inverting} is read as \textit{back-propagation} through the given deep neural network.} a Deep neural network.
    
  The key benefit of MIND lies in the fact that the deep convolutional network for frame-interpolation can be trained from ordinary video sequences without any man-made ground truth signals. The training data in our case is given by triplets of images, each one consisting of two input images and one output image that represents the ground-truth interpolated frame. A correct example of a ground truth output image is an image that---when inserted in between the input pair of images---forms a \textit{temporally coherent} sequence of frames. Such temporal coherency is naturally contained in regular video sequences, which allows us to simply use triplets of sequential images from almost arbitrary video streams for training our network. The first and the third frame of each triplet are used as inputs to the network, and the second frame as the ground truth interpolated frame. Most importantly, since the inversion of our network returns frame-to-frame correspondences, it therefore learns how to do image matching without any requirement for manually designed models or expensive ground truth correspondences. In other words, the presented approach learns image matching by simply ``watching videos''.
  
	The paper is organized as follows. Section \ref{sec:related_work} reviews relevant prior work. Section \ref{sec:methodology} explains the present \textit{analysis-by-synthesis} approach, including both the \textit{analysis} part of how MIND works and the \textit{synthesis} part of the deep convolutional architecture for frame interpolation. Section \ref{sec:experiment} demonstrates the surprising performance for the present purely unsupervised learning approach, which is comparable to several traditional empirically designed methods. Section \ref{sec:discussion} finally discusses our contribution and provides an outlook onto future works. 

\section{Related Work} 
\label{sec:related_work}
		
	\noindent\textbf{Deep learning meets image matching} Image matching is a classical problem in computer vision. Here we limit the discussion to recent works that address image matching through learning based approaches. Roughly speaking, there exist two lines of research for this topic: the first one consists of making use of features or representations learned by deep neural networks, which are either originally trained for other tasks such as object recognition \cite{long2014convnets,fischer2014descriptor}, or specially designed and trained for the purpose of image matching \cite{huang2012learning,agrawal2015learning,simo15}. The second major line of research employs deep neural networks to compute the similarity between image patches \cite{vzbontar2014computing,park2015leveraging,zagoruyko15}. In contrast to our work, the cited contributions mainly address sub-modules of image matching (feature extraction or matching cost computation), rather than providing end-to-end solutions. An exception is given by FlowNet \cite{fischer2015flownet}, which presents an interesting deep learning based approach for dense optic flow computation. It does however depend on ground truth flow for training the network. 
	
	It is also worth to mention that the Gated restricted Boltzmann machine model proposed by Memisevic and Hinton \cite{memisevic2007unsupervised} and then extended by Taylor et al. \cite{taylor2010convolutional} could also be trained in an unsupervised manner and be applied to infer constrained image transforms such as flow fields for ``shifting pixel''. However, this line of work is mainly aiming at learning motion features for understanding video data. It is similar to the works of temporal coherence learning mentioned below.
	
	\vspace{+0.1cm}
	\noindent\textbf{Temporal coherence learning} Unsupervised learning is a broad topic in the field of machine learning. Our discussion here focuses on works that exploit temporal coherency in natural videos, sometimes also called \textit{temporal coherence learning} \cite{mobahi2009deep,wiskott2002slow,becker1997learning}. As a recent representative work, Wang et al. \cite{wang2015unsupervised} exploit temporal coherency by visual tracking in videos, and report that the learned representation achieves competitive performance compared to some supervised alternatives. While temporal coherence learning mostly aims at learning features or representations, some recent works on reconstructing and predicting video frames in an unsupervised setting \cite{ranzato2014video} are closely related to our work as well. Srivastava et al. \cite{srivastava2015unsupervised} use an encoder LSTM to map input sequences into a fixed length representation, and use the latter for reconstructing the input or even predicting future frames. Goroshin et al. \cite{goroshin2015learning} consider videos as one-dimensional, time-parametrized trajectories embedded in a low dimensional manifold. They train deep feature hierarchies that linearise the transformations observed in natural video sequences for the purpose of frame prediction. Though related to our work, these works are not aiming at image matching. It will be interesting to apply our concept of matching by inverting to the above models for temporal coherence learning. 
	
	\vspace{+0.1cm}
	\noindent\textbf{Network inversion} Note that inverting a learned network is traditionally defined as reconstructing the input from the output of an artificial neural network \cite{jensen1999inversion}. Mahendran et al. \cite{mahendran2014understanding} and Dosovitskiy et al. \cite{dosovitskiy2015inverting} apply this concept to understand what information is preserved by a network. In our context, \textit{inverting a network} means \textit{back-propogation through a learned network in order to obtain the gradient map with respect to the input signals}. Interestingly, the idea has already been introduced in the work of Simonyan et al. \cite{simonyan2013deep}, emphasizing that the retrieved sensitivity maps may serve to identify image-specific class saliency. Similarly, Bach et al. \cite{bach2015pixel} employ gradient maps as a measure for the contribution of single pixels to nonlinear classifier, thus helping to explain how decisions are made.
	
	\section{Methodology}
	\label{sec:methodology}
	
	The \textit{analysis by synthesis} approach for dense image matching is described in this section: we first explain the \textit{analysis} part, i.e. how to obtained correspondences given the trained neural network and the interpolated image. For the \textit{synthesis} part, it is described here the detailed architecture of the deep convolutional network designed for frame interpolation.
	
	\subsection{Matching by Inverting a Deep neural network}
	
	Assuming that we have a well trained deep neural network for frame interpolation in our hand, the core technical question behind our work is how to recover the correspondences between the input pair of images from there. As explained previously, dense correspondence matching may be regarded as a sub-problem of frame-interpolation, which is why we should be able to trace back the matches starting from the interpolated frame generated during the forward-propagation through the trained network. Our task then consists of back-tracking each pixel in the output image to exactly one pixel in each of the two input images. Note that this back-tracking does not mean reconstructing input images from the output one. Instead, we only need to find the pixels in each input image which have the maximum influence to each pixel of the output image.
	
	We perform back-tracking by applying a technique similar to the one adopted by Simonyan et al. \cite{simonyan2013deep}. For each pixel in the output image, we compute the gradient of its value with respect to each input pixel, thus telling us how much it is under the influence of individual pixels at the input. The gradient is computed based on back-propagation, and leads to sensitivity or influence maps at the input of the network. 
	
	From a more formal perspective, our approach may be explained as follows. Let $\mathbf{I}_{2} = \mathcal{F}(\mathbf{I}_{1},\mathbf{I}_{3})$ denote a non-linear function (i.e. the trained deep neural network) that describes the mapping from two input images $\mathbf{I}_{1}$ and $\mathbf{I}_{3}$ to an interpolated image $\mathbf{I}_{2}$ lying approximately at the ``center'' of the input frames. Thinking of $\mathcal{F}$ as a vectorial mapping, it can be split up into $h\times w$ non-linear sub-functions, each one producing the corresponding pixel in the output image
	\begin{equation}
		\mathcal{F}(\mathbf{I}_{1},\mathbf{I}_{3}) = \left ( \begin{matrix}
			f^{11}(\mathbf{I}_{1},\mathbf{I}_{3}) & \ldots & f^{1w}(\mathbf{I}_{1},\mathbf{I}_{3}) \\
			\vdots & & \vdots \\
			f^{h1}(\mathbf{I}_{1},\mathbf{I}_{3}) & \ldots & f^{hw}(\mathbf{I}_{1},\mathbf{I}_{3})
		\end{matrix} \right)_{h\times w}.
	\end{equation}
	In order to produce the sensitivity maps, we apply back-propagation to compute the Jacobian matrix with respect to each input image individually. The Jacobian with respect to the first image is given by
	\begin{equation}
		\frac{\partial \mathcal{F}(\mathbf{I}_{1},\mathbf{I}_{3})}{\partial \mathbf{I}_{1}} = \left ( \begin{matrix}
			\frac{\partial f^{11}(\mathbf{I}_{1},\mathbf{I}_{3})}{\partial \mathbf{I}_{1}} & \ldots & \frac{\partial f^{1w}(\mathbf{I}_{1},\mathbf{I}_{3})}{\partial \mathbf{I}_{1}} \\
			\vdots & & \vdots \\
			\frac{\partial f^{h1}(\mathbf{I}_{1},\mathbf{I}_{3})}{\partial \mathbf{I}_{1}} & \ldots & \frac{\partial f^{hw}(\mathbf{I}_{1},\mathbf{I}_{3})}{\partial \mathbf{I}_{1}}
		\end{matrix} \right)_{h \times h \times w \times w},
	\end{equation}
	illustrating that this derivative results in one $h\times w$ matrix for each one of the $h\times w$ pixels at the output. The Jacobian with respect to $\mathbf{I}_{3}$ is given in a similar way. Let's define the absolute gradients of the output point $(i,j)$ with respect to each one of the input images, and evaluated for the concrete inputs $\mathbf{\hat{I}}_{1}$ and $\mathbf{\hat{I}}_{3}$. They are given by
	\begin{equation}
		\left\{ \begin{matrix}
			\mathcal{G}^{i,j}_{\mathbf{I_{1}}}(\mathbf{\hat{I}}_{1},\mathbf{\hat{I}}_{3}) = \operatorname{abs} \left( \left. \frac{\partial f^{ij}(\mathbf{I}_{1},\mathbf{I}_{3})}{\partial \mathbf{I}_{1}} \right|_{ \tiny{\begin{matrix} \mathbf{I}_{1}=\mathbf{\hat{I}}_{1} \\ \mathbf{I}_{3}=\mathbf{\hat{I}}_{3} \end{matrix} } } \right) \\
			\mathcal{G}^{i,j}_{\mathbf{I_{3}}}(\mathbf{\hat{I}}_{1},\mathbf{\hat{I}}_{3}) = \operatorname{abs} \left( \left. \frac{\partial f^{ij}(\mathbf{I}_{1},\mathbf{I}_{3})}{\partial \mathbf{I}_{3}} \right|_{ \tiny{\begin{matrix} \mathbf{I}_{1}=\mathbf{\hat{I}}_{1} \\ \mathbf{I}_{3}=\mathbf{\hat{I}}_{3} \end{matrix} } } \right)
		\end{matrix} \right. ,
	\end{equation}
	where $\operatorname{abs}$ replaces each entry of a matrix by its absolute value. The gradient maps produced in this way notably represent the seeked sensitivity or influence maps that may now serve in order to derive the coordinates of each correspondence. We notably extract the most responsible point in each gradient map, and connect those two points in order to return the correspondence.
	
	In the spirit of unsupervised learning, we opted for the simplest possible choice of taking the coordinates of the maximum entry in $\mathcal{G}^{i,j}_{\mathbf{I_{1}}}(\mathbf{\hat{I}}_{1},\mathbf{\hat{I}}_{3})$ and $ \mathcal{G}^{i,j}_{\mathbf{I_{3}}}(\mathbf{\hat{I}}_{1},\mathbf{\hat{I}}_{3})$, respectively. Let us denote these points with $c^{ij}_{\mathbf{I}_{1}}$ and $c^{ij}_{\mathbf{I}_{3}}$. By computing the two gradient maps for each point in the output image and extracting each time the most responsible point, we thus obtain the following two lists of points
	\begin{equation}
		\left\{ \begin{matrix}
			\mathcal{C}_{\mathbf{I}_{1}} = \left\{  c^{ij}_{\mathbf{I}_{1}} \right\}\\
			\mathcal{C}_{\mathbf{I}_{3}} = \left\{  c^{ij}_{\mathbf{I}_{3}} \right\}
		\end{matrix}\right. , i=1,\ldots,h, j=1,\ldots,w
	\end{equation}
	The set of correspondences $\mathcal{S}$ is then given by combining same-index elements from $\mathcal{C}_{\mathbf{I}_{1}}$ and $\mathcal{C}_{\mathbf{I}_{3}}$, eventually resulting in
	\begin{eqnarray}
		\mathcal{S} & = & \left\{ s^{ij} \right\}, i=1,\ldots,h, j=1,\ldots,w \nonumber \\
		& = & \left\{ \left\{c^{11}_{\mathbf{I}_{1}}, c^{11}_{\mathbf{I}_{3}} \right\}, \ldots, \left\{c^{hw}_{\mathbf{I}_{1}}, c^{hw}_{\mathbf{I}_{3}} \right\} \right\}.
	\end{eqnarray}			
				    
	\subsection{Deep neural network for Frame Interpolation}	
	\label{sec:architecture}	
		
\begin{figure}[t]
			\centering
			\includegraphics[width = 0.43\textwidth]{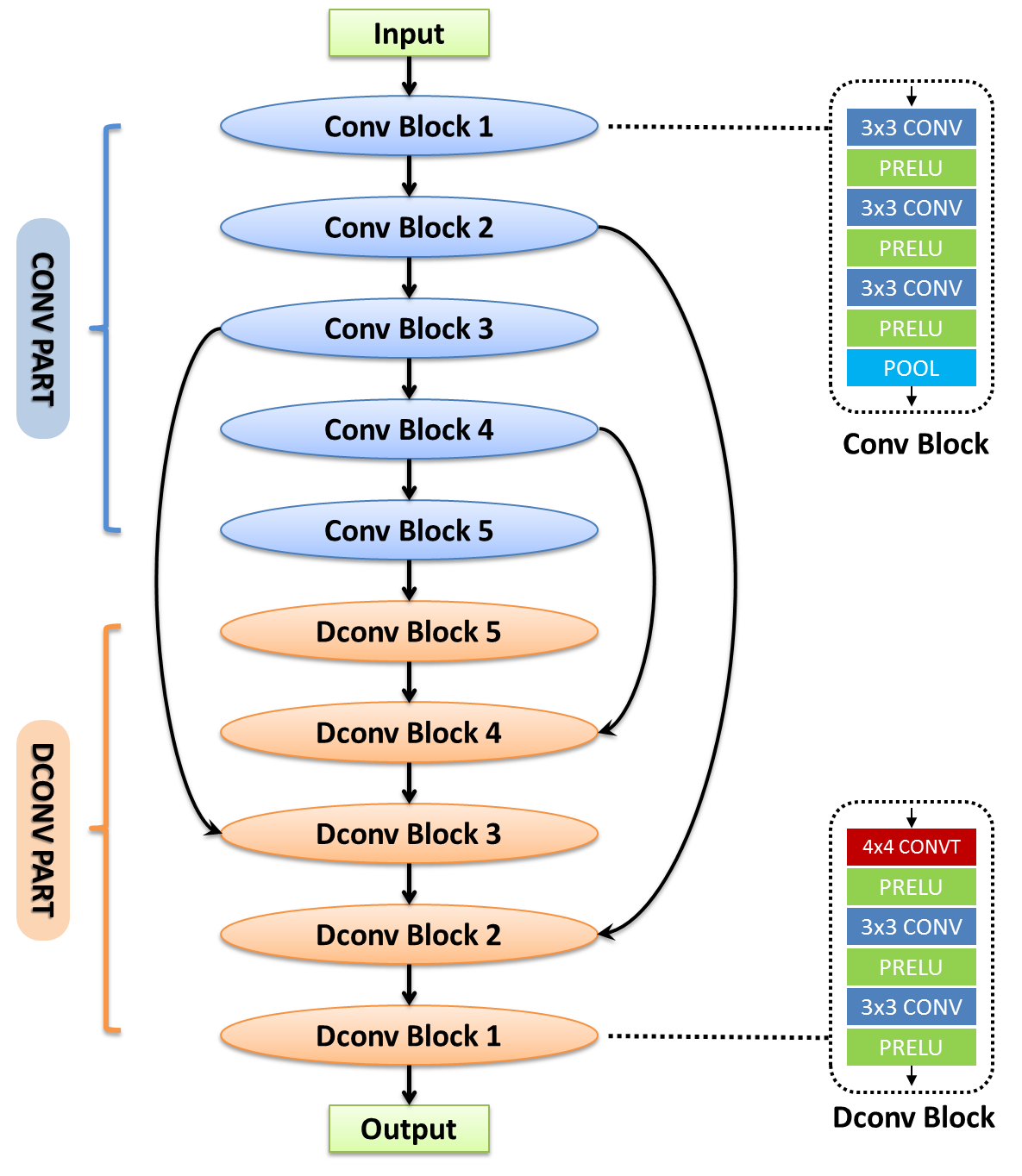}
			\caption{Architecture of our network. The network takes 2 RGB images as an input to produce the interpolated RGB image. Please note that Dconv Block 4 takes the outputs from both Conv Block 2 and Dconv Block 5 as input. Dconv Block 3 and Dconv Block 2 have a similar input configuration.}
			\label{fig:architecture}
		\end{figure}

	The architecture of our frame-interpolation network is inspired by \textit{FlowNetSimple} as presented in Fischer et al. \cite{fischer2015flownet}. As illustrated in Figure \ref{fig:architecture}, it consists of a Convolutional Part and a Deconvolutional Part. The two parts serve as ``encoder'' and ``decoder'' respectively, similar to the auto-encoder architecture presented by Hinton and Salakhutdinov \cite{hinton2006reducing}. The basic block within the Convolutional Part---denoted Convolution Block---follows the common pattern of the convolutional neural network architecture: 
		\begin{spacing}{1.3}
		\centerline{\fontsize{7}{10}\textbf{INPUT --\textgreater [CONV --\textgreater PRELU] * 3 --\textgreater POOL --\textgreater OUTPUT.}}
		\end{spacing}
        \noindent The Parametric Rectified Linear Unit \cite{he2015delving} is adopted in our work. Following the suggestions from VGG-Net \cite{chatfield2014return}, we set the size of the receptive field of all convolution filters to three---along with a stride and a padding of one---and duplicate {\fontsize{7}{10}\textbf{[CONV --\textgreater PRELU]}} three times to better model the non-linearity. 
		
		The Deconvolution Part consists of Deconvolution Blocks, each one including a convolution transpose layer \cite{vedaldi2014matconvnet} and two convolution layers. The first one has a receptive field of four, a stride of two, and a padding of one. The pattern of the Deconvolution Block follows:
		\begin{spacing}{1.3}
		\centerline{\fontsize{7}{10}\textbf{INPUT --\textgreater [CONVT --\textgreater PRELU] --\textgreater [CONV --\textgreater PRELU] * 2 --\textgreater OUTPUT.}}
		\end{spacing}
		\noindent In order to maintain fine-grained image details in the interpolation frame, we make a copy of the output features produced by Convolution Blocks 2, 3, and 4, and concat them as an additional input to the Deconvolution Blocks 4, 3, and 2, respectively. This concept is illustrated by the side arrows in Figure \ref{fig:architecture}, and similar ideas have already been used in prior work \cite{fischer2015flownet,eigen2014depth}. Recent works \cite {srivastava2015highway,he2015deep} indicate that the `side arrows' may also help to better train the deep network.
		
		It is easy to notice that our network is a fully convolutional one, thus allowing us to feed it with images of different resolutions. This is an important advantage, as different datasets may use different height-to-width ratios. The output blob size for each block in our network is listed in Table \ref{tab:cnn_size}.
	
		\begin{table*}
		\caption{The table lists the output blob size of each block in our network. Note that we stack two RGB images into one input blob, and thus the depth is 6. The output of the network is an RGB image and thus the depth equals to 3. The indicated widths are for the network trained on KITTI. The ones for the Sintel data are easily obtained, the only difference being that the input images are scaled to 256$\times$128 rather than 384$\times$128.}
			\centering
			\begin{center}
			\scalebox{0.7}{
				\begin{tabular}{|c |c |c |c |c |c | c| c| c| c| c| c| c|}
					\hline
					& {Input} & {Conv1} & {Conv2} & {Conv3} & {Conv4} & {Conv5} &{Dconv5} & {Dconv4} & {Dconv3} & {Dconv2} & {Dconv1} & {Output}\\
					\hline
					{Depth}  & 6 & 96 & 96 & 128 & 128 & 128 & 128 & 128 & 128 & 96 & 96 & 3\\
					\hline
					{Height} & 128 & 64  & 32 & 16 & 8  & 4  & 8  & 16 & 32 & 64  & 128 & 128\\
					\hline
					{Width}  & 384 & 192 & 96 & 48 & 24 & 12 & 24 & 48 & 96 & 192 & 384 & 384\\
					\hline
				\end{tabular}
			} 
			\end{center}
			\label{tab:cnn_size}
		\end{table*}
		
	\section{Experiments}
	\label{sec:experiment}
	
	In this section, we first explain the implementation details behind MIND such as training data and loss function. The examples as proofs of concept for MIND are introduced before a discussion on the generalization ability of the trained CNN. We finally evaluate MIND in terms of quantitative matching performance and compare it to traditional image matching methods. 
	
	\vspace{-0.1cm}
	\subsection{Implementation Details}
	
	\noindent\textbf{Training Data:} Quantity and quality of training data are crucial for training a deep neural network. However, our case is particularly easy as we can simply use huge amounts of real-world videos. In this work, we focus on training with the KITTI RAW videos \cite{Geiger2013IJRR} and Sintel videos\footnote{Sintel, the Durian Open Movie Project. https://durian.blender.org/} and show that the resulting learned network performs reasonably well. The network is first trained with the KITTI RAW video sequences which are captured by driving around the city of Karlsruhe, through rural areas and over highways. The dataset contains 56 image sequences with in total 16,951 frames. For each sequence, we take every three consecutive frames (both in forward and backward direction) as a training triplet, where the first and the third image serve as inputs to the network and second image as the corresponding output. These images are then augmented by vertical flipping, horizontal flipping and a combination of both. The total number of sample triplets is 133,921. We then fine-tune the network on examples selected from the original Sintel movie. We manually collected 63 video clips with in total 5,670 frames from the movie. After grouping and data augmentation we finally obtain 44,352 sample triplets. Note that, compared to the KITTI sequences which are recorded with relatively uniform velocity, the Sintel sequences represent more difficult training examples in the context of our work, as they contain a lot of fast and unrealistic motion captured with a frame rate of only 24 fps. A significant portion of the Sintel samples therefore does not contain the required temporal coherence. We will discuss this issue further in Section \ref{sec:examples}.
		
	\vspace{+0.1cm}
	\noindent\textbf{Loss Function:} Several previous works \cite{goroshin2015learning,wang2015unsupervised} mention that minimizing the L2 loss between the output frame and the training example may lead to unrealistic and blurry predictions. We have not been able to confirm this throughout our experiments, but found that the Charbonnier loss $\rho(x)=\sqrt{(x^2 + \epsilon ^2)}$ commonly employed for robust optical flow computation \cite{sun2014quantitative} leads to an improvement over the L2 loss. We employ it to train our network, with $\epsilon$ set to 0.1.
	
	\vspace{+0.1cm}			
	\noindent\textbf{Training Details:} The training is performed using Caffe \cite{jia2014caffe} on a machine with two K40c GPUs. The weights of the network are initialized by Xavier's approach \cite{glorot2010understanding} and optimized by the Adam solver \cite{kingma2014adam} with a fixed momentum of 0.9. The initial learning rate is set to 1e-3 and then manually tuned down once ceasing of loss reduction sets in. For training on the KITTI RAW data, the images are scaled to 384$\times$128. For training on the Sintel dataset, the images are scaled to 256$\times$128. The batch size is 16. We run the training on KITTI RAW from scratch for about 20 epochs, and then fine-tuned it on the Sintel movie images for 15 epochs. We did not observe over-fitting during training, and terminated the training after 5 days.
	
	\vspace{+0.1cm}
	\noindent\textbf{Execution time:} MIND can be applied to different scenarios (e.g. sparse or dense matching). We focus here on semi-dense image matching in order to obtain a result comparable with other methods. We compute the correspondences across the input images for each corner of a predefined raster grid of 4 pixels width in the interpolated image. Note that MIND currently depends on a large amount of computational resources as it performs back-propagation through the entire network for every pixel that needs to be matched. For an image of size 384$\times$128, each forward pass through our network takes 40ms on a PC with K40c GPU, and each backward pass takes 158ms. For each image pair, we need to perform one forward pass to first obtain the interpolation. We then need to perform 384$\times$128 / 4 / 4 = 3072 backward passes to find the correspondences, resulting in a total of about 486 seconds (8 minutes).

	\subsection{Qualitative examples for Interpolation and Matching}
	\label{sec:examples}
	
	We demonstrate here the visual examples as proofs of concept for how the present approach works on both tasks of frame interpolation and image matching. We further introduce a discussion on the generalization ability of the trained model.
	
	\begin{figure*}[t]
		\centering
		\includegraphics[width = 1.0\textwidth]{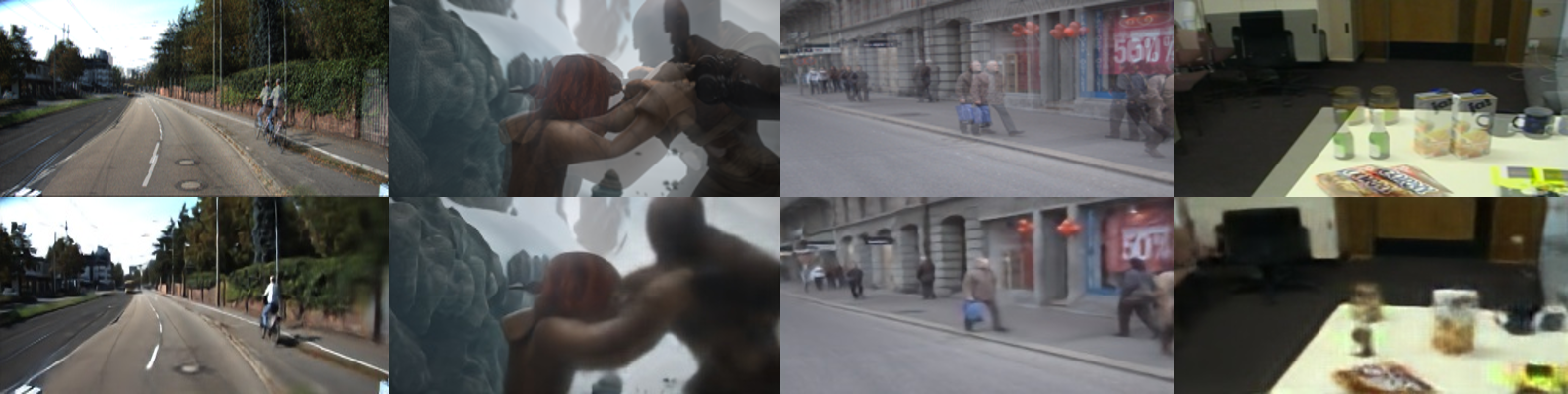}
		\caption{Examples of Frame interpolation (best viewed in colour). From left to right: example on KITTI, Sintel, ETH Multi-Person Tracking dataset\cite{ess2009robust} and Bonn Benchmark on Tracking \cite{KleinIROS10}, respectively. In each column, the first image is an overlay of the two input frames. The second one is the interpolated image obtained by our network. For the first example, we use the network trained on KITTI itself. For all others, we use the network fine-tuned on Sintel data.}
		\label{fig:Interpolation_examples}
	\end{figure*}	
	
	\vspace{+0.1cm}
	\noindent\textbf{Examples for frame interpolation:} We show the examples of frame interpolation in Figure \ref{fig:Interpolation_examples}. The first two columns show the examples on KITTI and Sintel images which are taken from the validation datasets originally collected for the purpose of monitoring the network training process. It can be seen that the trained CNNs cover the motion correctly for both KITTI and Sintel image pairs. It could be noticed as well that some fine-grained details are not preserved well in both examples, even though we have put special considerations when designing the convolutional architecture, c.f. section \ref{sec:architecture}. Nevertheless, we would like to mind the readers that the goal of the present work is not to provide a state-of-the-art frame interpolation algorithm. And for the goal of image matching, we will see that the preservation of perfect image details is in fact not necessary. 

\begin{figure*}[t]
		\centering
		\includegraphics[width = 1.0\textwidth]{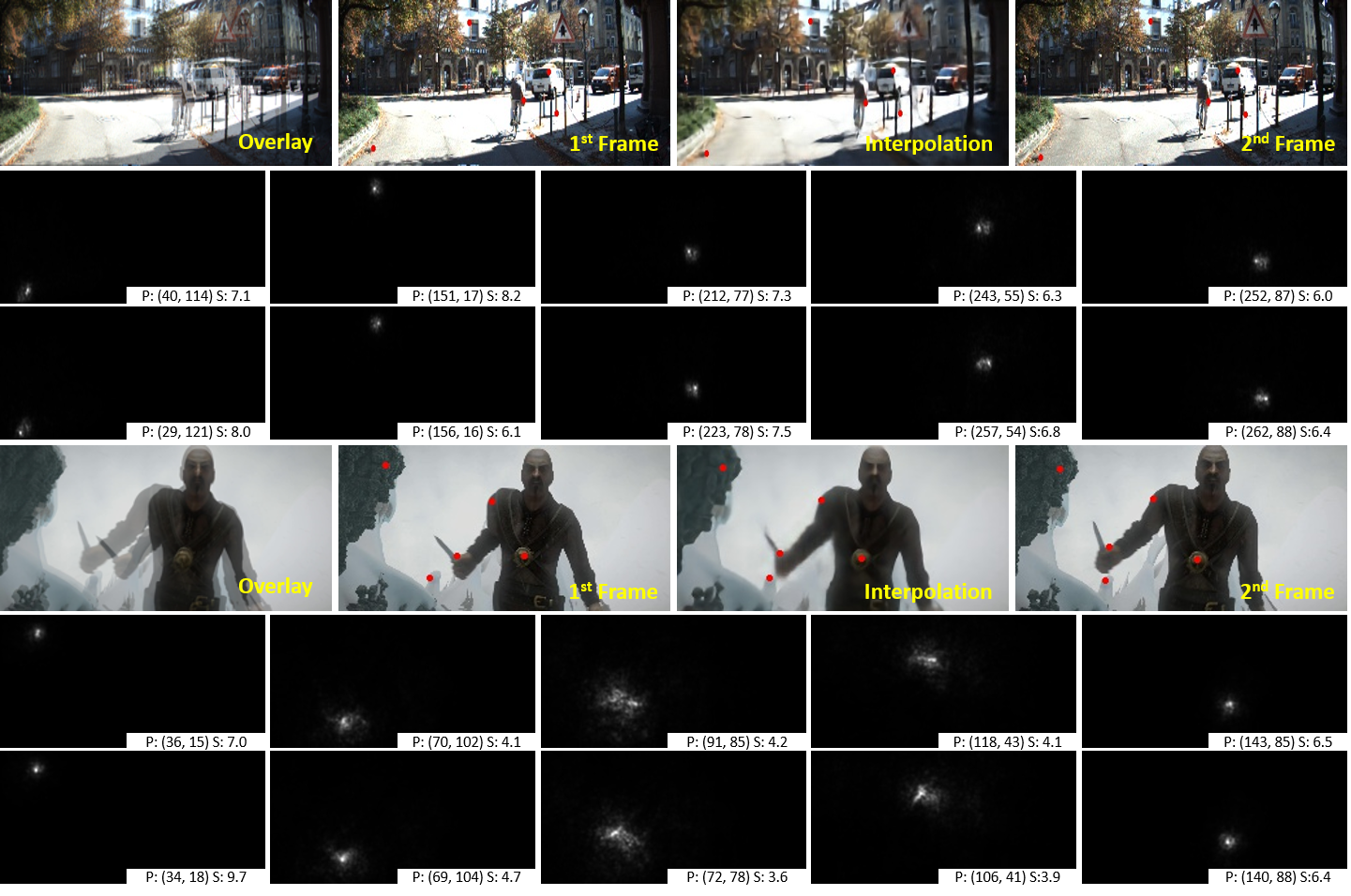}
		\caption{Two matching examples for image pairs taken from the KITTI RAW video and the Sintel movie clip (best viewed in colour). For each example, the corresponding row of images shows input image 1, the interpolated image, and then input image 2 (from left to right). The red points mark five sample correspondences. The two rows below each example show the gradient/saliency maps for each match (from left to right) in each input image (maps for input image 1 on top, and maps for input image 2 in the bottom). The figures also indicate the coordinates of the maximal gradient location (P) along with the corresponding matching score (S). The matching score is defined as the ratio between the maximum gradient intensity and the mean gradient intensity within a 20$\times$20 area around P.}
		\label{fig:Matching_examples}
	\end{figure*}

	\vspace{+0.1cm}
	\noindent\textbf{Examples for image matching:} Here we present examples to demonstrate how MIND obtains correspondences given the trained CNNs for frame interpolation. The examples taken from KITTI and Sintel videos are shown in Figure \ref{fig:Matching_examples}. By computing the gradient of manually marked pixels in the interpolated image, MIND successfully obtains correct correspondences between the 2 input images. It can be seen that the correct correspondences are obtained even in some fast moving areas where fine-grained image details are missed, e.g. the area of the character's shaking hand in the Sintel example.
	
\begin{figure*}[!t]
		\centering
		\includegraphics[width = 1.0\textwidth]{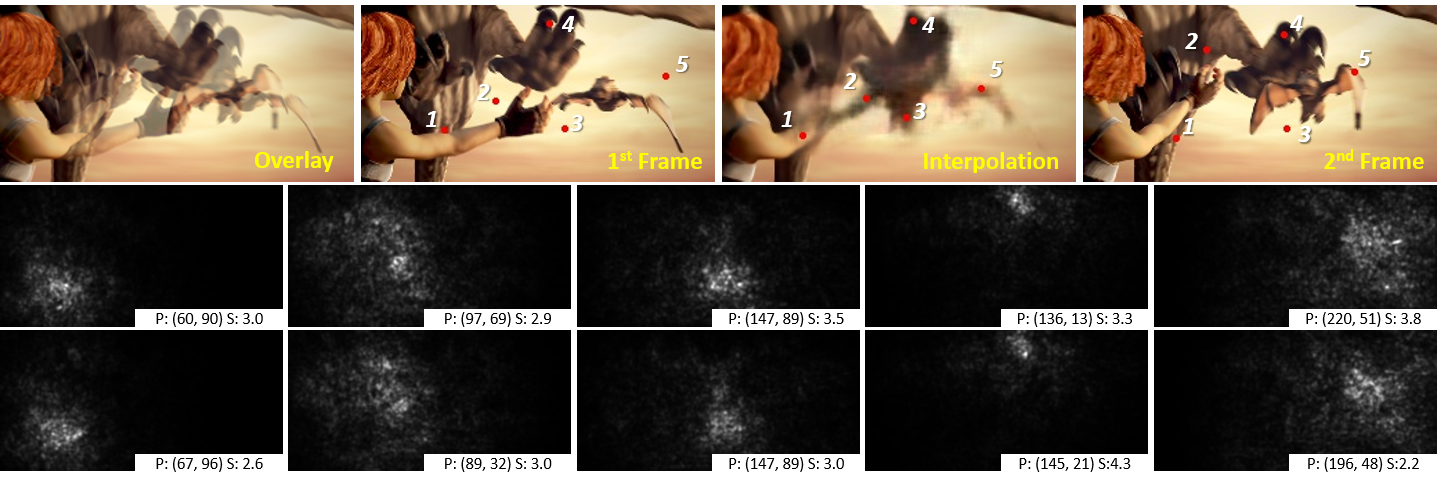}
		\caption{Failure example of MIND for image pair taken from the Sintel movie clip (best viewed in colour). The gradient/saliency maps (from left to right) are for matches labelled as 1, 2, \dots, 5,  respectively. }
		\label{fig:Matching_examples_fail}
	\end{figure*}

	We further show one failure example taken from Sintel images. In Figure \ref{fig:Matching_examples_fail}, it can be observed that the interpolation fails as the motion of the small dragon and the character's hand have not been covered correctly. It then comes as no surprise that MIND fails to extract correct matches for almost all of the selected points. However, it is worth to note that the No.4 match has better quality than others, of which the corresponding gradient maps are less distinctive. The matching score/confidence returned by MIND is inspired by this behaviour and defined as the ratio between the maximum gradient intensity and the mean gradient intensity within a small area around the maximal gradient location.

	As illustrated in Section \ref{sec:matching}, the general performance of MIND, especially on KITTI images, is good. The failure example in Figure \ref{fig:Matching_examples_fail} outlines a extreme case in the Sintel sequences dominated by fast and highly non-rigid motion in the scene.

	\vspace{+0.1cm}
	\noindent\textbf{Generalization ability:} It is essential for learning based approaches to hold good generalization ability. Though MIND enjoys the benefit that it can learn image matching by just ``watching videos'' (i.e. it could first do fine-tuning in the given image sequences and perform the interpolation \& matching after that), it is important to verify whether the present CNN is indeed learning the ability to interpolate frames and match images, rather than only ``remember'' the KITTI or Sintel-like images. 
	
	We demonstrate the generalization ability of the trained CNN by applying it to images taken from the ETH Multi-Person Tracking dataset \cite{ess2009robust} and the Bonn Benchmark on Tracking \cite{KleinIROS10}, which have not been used for either training or fine-tuning. The results are showed in Figure \ref{fig:Interpolation_examples}, from which we can see that the trained CNN again covers the motion correctly. It provides evidence about what has been learned by ``watching videos''.
	
	\begin{figure}[t]
		\centering
		\includegraphics[width = 1.0\textwidth]{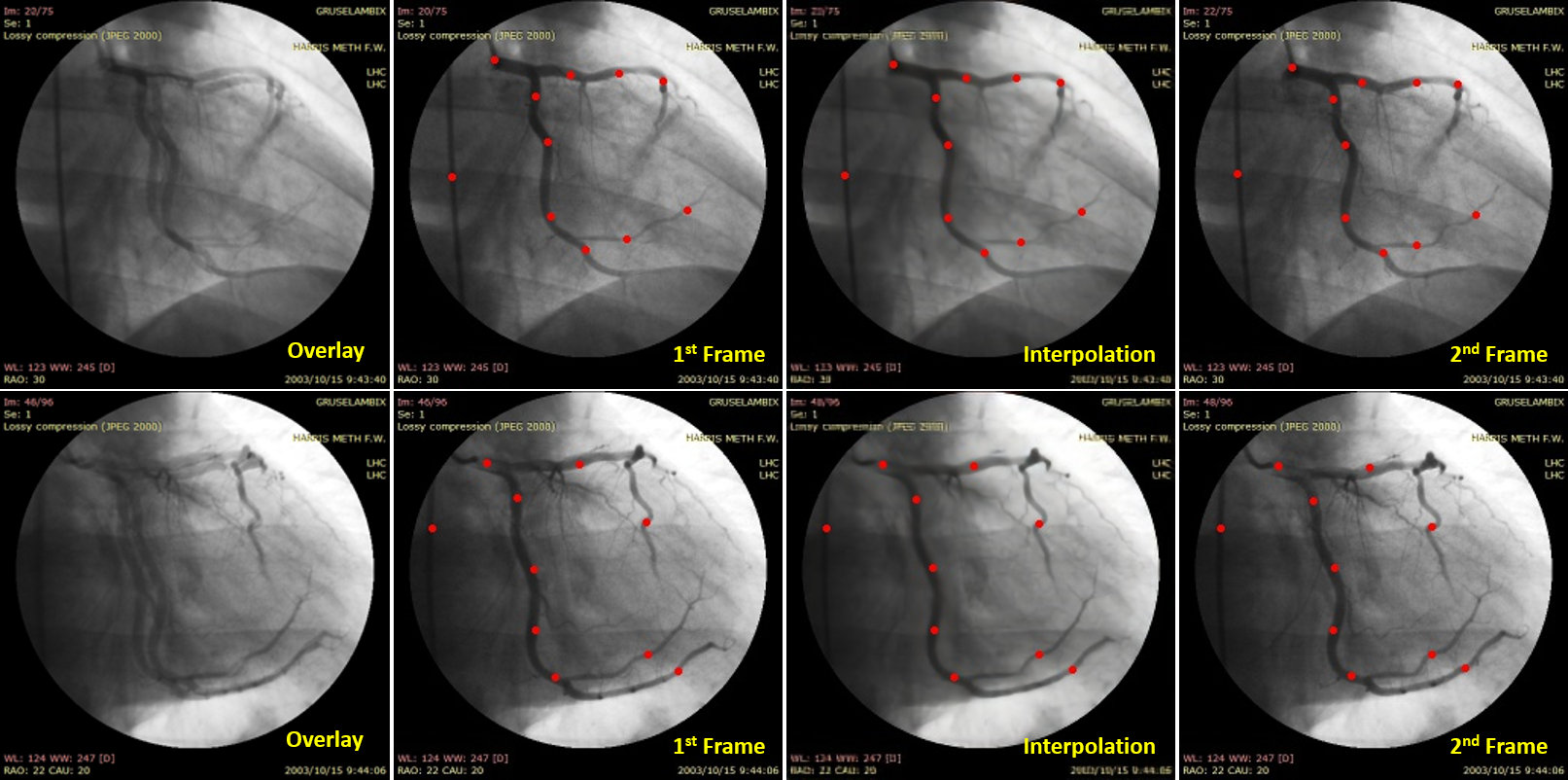}
		\caption{Examples of MIND on DICOM images. There are two examples shown in different rows. For each example,  the overlay of input image-pair, 1st input image, interpolation returned by the CNN and the 2nd input image are shown in the columns from left to right, respectively. The red points in columns 2, 3 and 4 indicate the matches obtained by MIND.}
		\label{fig:Interpolation_dcm}
	\end{figure}
	
	The generalization ability is further illustrated by applying MIND to DICOM images of coronary angiogram\footnote{The images are taken from a DICOM sample image set: http://www.osirix-viewer.com/datasets/. Alias Name: GRUSELAMBIX.}. In Figure \ref{fig:Interpolation_dcm}, it can be seen that these images are substantially different from natural ones. Though again failing to preserve perfect image details, the CNN, which is trained on natural images, performs impressively well on the DICOM images. The nice generalization ability of the CNN is underlined by results on both frame interpolation and image matching.

	\vspace{-0.2cm}
	\subsection{Quantitative Performance of Image Matching}
	\label{sec:matching}
	\vspace{-0.3cm}

        \begin{table}
		\caption{Matching performance on the KITTI 2012 flow training set. DeepM denotes DeepMatching. Metrics: Average Point Error (APE) (the lower the better), and Accuracy@T (the higher the better).  Bold numbers indicate best performance, underlined numbers 2nd best.}
		\begin{center}
		\scalebox{0.7}{
			\begin{tabular}{c|c  c  c  c}
				\hline
				& {\hspace{0.2cm}MIND\hspace{0.2cm}} &{\hspace{0.2cm}DeepM\hspace{0.2cm}} & {\hspace{0.2cm}HoG\hspace{0.2cm}} & {\hspace{0.2cm}KLT\hspace{0.2cm}}\\
				\hline
				{APE}  & \underline{4.695} & \textbf{3.442} & 9.680 & 8.157 \\
				\hline
				{Accuracy@5}  & \underline{0.716} & \textbf{0.835} & 0.455 & 0.702 \\
				\hline
				{Accuracy@10} & \underline{0.915} & \textbf{0.953} & 0.805 & 0.826 \\
				\hline
				{Accuracy@20} & \underline{0.981} & \textbf{0.987} & 0.929 & 0.903 \\
				\hline
				{Accuracy@30} & \textbf{0.993}    & \textbf{0.993} & 0.959 & 0.938 \\
				\hline
			\end{tabular}}
		\end{center}
		\label{tab:Matching_KITTI}
	\end{table}	
	
	\begin{table}
		\footnotesize
		\caption{Matching performance on the MPI-Sintel training set (Final pass). DeepM denotes DeepMatching. Metrics: Average Point Error (APE) (the lower the better), and Accuracy@T (the higher the better). Bold numbers indicate best performance, underlined numbers 2nd best.}
		\begin{center}
		\scalebox{0.7}{
			\begin{tabular}{c|c c c c}
				\hline
				& {\hspace{0.2cm}MIND\hspace{0.2cm}} &{\hspace{0.2cm}DeepM\hspace{0.2cm}} & {\hspace{0.2cm}HoG\hspace{0.2cm}} & {\hspace{0.2cm}KLT\hspace{0.2cm}}\\
				\hline
				{APE}  & \underline{5.838} & \textbf{3.240} & 7.856 & 8.836 \\
				\hline
				{Accuracy@5}  & 0.719 & \textbf{0.875} & 0.688 & \underline{0.808} \\
				\hline
				{Accuracy@10}  & \underline{0.876} & \textbf{0.951} & 0.875 & 0.864 \\
				\hline
				{Accuracy@20}  & \underline{0.948} & \textbf{0.977} & 0.947 & 0.906 \\
				\hline
				{Accuracy@30}  & \underline{0.967} & \textbf{0.986} & 0.964 & 0.927 \\
				\hline
			\end{tabular}}
		\end{center}
		\label{tab:Matching_Sintel}
	\end{table}		
	
	We compare the matches produced by MIND against those of several empirically designed methods: the classical Kanade--Lucas--Tomasi feature tracker \cite{bouguet2001pyramidal}, HoG descriptor matching \cite{brox11} (which is widely employed to boost dense optical flow computation), and the more recent DeepMatching approach \cite{weinzaepfel2013deepflow} which relies on a multilayer convolutional architecture and achieves state-of-the-art performance. As observed in \cite{weinzaepfel2013deepflow}, comparing different matching algorithms is delicate because they usually produce different numbers of matches for different parts of the image. For the sake of a fair comparison, we adjust the parameters of each algorithm to make them produce as many as possible matches with an as homogeneous as possible distribution across the input images. For DeepMatching, we use the default parameters. For MIND, we extract correspondences for each corner of a uniform grid of 4 pixels width. For KLT, we set the minEigThreshold to 1e-9 to generate as many matches as possible. For HoG, we again set the pixel sampling grid width to 4. We then sort the matches according to suitable metrics\footnote{For DeepMatching, we sort the matches according to the matching score given by the open source code \cite{weinzaepfel2013deepflow}. For KLT, the metric is the error returned by the OpenCV implementation \cite{bradski2008learning}. For HoG, we use the matching score defined in \cite{brox11}. For MIND, the matching score is defined as the ratio between the maximum gradient intensity and the mean gradient intensity within a 20$\times$20 area around the maximal gradient location.} and select the same amount of ``best'' matches for each algorithm. In this way, the 4 algorithms produce the same numbers of matches with similar coverage over each input image.
	
	The comparisons are performed on both KITTI \cite{Geiger2013IJRR} and MPI-Sintel \cite{Butler:ECCV:2012} training sets where ground truth correspondences can be extracted from the available ground truth flow fields. We perform all of our experiments on the same image resolution than the one used by our network.\footnote{It is ideal to evaluate both image matching and optical flow in benchmarks of KITTI and MPI-Sintel. Due to the fact that the present MIND is currently designed only for resolution-reduced images, we can't process the benchmark datasets directly, but apply all algorithms locally to the test datasets, followed by the standard evaluation and error metrics known from prior art.} On KITTI, the images are scaled to 384$\times$128, and for MPI-Sintel, 256$\times$128. We use the network trained on the KITTI RAW sequences for the matching experiment on the KITTI Flow 2012 training set. We then use the network fine-tuned on Sintel movie clips for the experiments on the MPI-Sintel Flow training set. The 4 algorithms are evaluated in terms of the Average Point Error (APE) and the Accuracy@T. The latter is defined as the proportion of ``correct'' matches from the first image with respect to the total number of matches \cite{revaud2015deep}. A match is considered correct if its pixel match in the second image is closer than T pixels to ground-truth.
	
	As can be observed in Table \ref{tab:Matching_KITTI} and Table \ref{tab:Matching_Sintel}, DeepMatching produces matches with the highest quality in terms of all metrics and on both MPI-Sintel and KITTI sets. Notably, MIND performs very close to DeepMatching on KITTI and outperforms KLT tracking and HoG matching by a considerable amount in terms of Accuracy@10 and Accuracy@20. It is surprising to see that MIND---an unsupervised learning based approach---works so well. The performance on MPI-Sintel however drops a bit due to the difficulty of the contained unrealistic motion. Though the APE measure indicates better performance than HoG and KLT, it is only safe to conclude that MIND remains competitive in terms of overall performance on MPI-Sintel, which can be seen further in the next section.
			    		    
	\vspace{-0.4cm}
	\subsection{Ability to Initialise Optical Flow Computation}
	
	\vspace{-0.3cm}

	To further understand the matching quality produced by MIND, we replace the DeepMatching part of DeepFlow\cite{weinzaepfel2013deepflow} with MIND to see whether MIND matches are able to boost optical flow performance in a similar way than DeepMatching and HoG or KLT matches. Similar to the evaluation in \cite{weinzaepfel2013deepflow}, we feed DeepFlow with matches obtained by each matching method in the previous section. The parameters (e.g. the matching weight) of DeepFlow are tuned accordingly to make best use of the pre-obtained matches. Note that we scale down the input images to 384$\times$128 for KITTI and 256$\times$128 for MPI-Sintel. We then up-size the obtained flow field to the original resolution by bilinear interpolation, to the end of comparing results in full resolution.
	
	The results on the KITTI Flow 2012 training set are indicated in Table \ref{tab:Flow_KITTI}. It can be seen that using the matches obtained by any of the 4 algorithms improves the flow performance compared to the case where we use no matches for initialization. Notably, MIND again reaches closest performance to DeepMatching in terms of all metrics, thus underlining the good matching quality obtained by MIND (better than KLT and HoG and comparable to DeepMatching). Table \ref{tab:Flow_Sintel} shows the results obtained on the MPI-Sintel training dataset. As in KITTI, the pre-obtained matches indeed help to improve the optical flow results especially in terms of the APE and s40+ metrics, while flow initialized by DeepMatching remains best overall. The results initialized from MIND matches however rank behind those initialized by HoG or KLT matches, which again suggests the importance of temporal coherency for training our network. The reason why KLT works better than in the evaluation presented in \cite{weinzaepfel2013deepflow} is because we run KLT in the downscaled images rather than the full resolution ones, and this helps KLT to better deal with large displacements.
	
	From the quantitative evaluations of matching and flow performance, it should be concluded that MIND works well on the KITTI Flow training set and achieves comparable performance to the state-of-the-art defined by DeepMatching. In the MPI-Sintel Flow training set, MIND still obtains comparable performance to the traditional HoG and KLT methods. The latter should still be interpreted as a good result especially considering that the quality of training data for the unrealistic Sintel images is insufficient.

	\begin{table}[!t]
		\caption{Flow performance on KITTI 2012 flow training set (non-occluded areas). out-\textit{x} refers to the percentage of pixels where flow estimation has an error above \textit{x} pixels.}
		\begin{center}
		\scalebox{0.7}{
			\begin{tabular}{c|c c c c c}
				\hline
				& {MIND} &{DeepM} & {HoG} & {KLT} & {No match}\\
				\hline
				{APE}   & \underline{2.89}   & \textbf{2.63}   & 3.06   & 3.40   & 3.55\\
				\hline
				{out-2}  &\underline{17.70\%}   &\textbf{17.09\%}   &17.89\%   &18.34\%   &18.49\%\\
				\hline
				{out-5}  & \underline{9.86\%}   & \textbf{9.18\%}   &10.05\%   &10.58\%   &10.77\%\\
				\hline
				{out-10} & \underline{6.45\%}   & \textbf{5.84\%}   & 6.66\%   & 7.20\%   & 7.40\%\\
				\hline
			\end{tabular}}
		\end{center}
		\label{tab:Flow_KITTI}
	\end{table}	
	
	\begin{table}[!t]
		\caption{Flow performance on MPI-Sintel flow training set. s0-10 is the APE for pixels with motions between 0 and 10 pixels. Similarly for s10-40 and s40+. }
		\begin{center}
		\scalebox{0.7}{
			\begin{tabular}{c|c c c c c}
				\hline
				& {MIND} &{DeepM} & {HoG} & {KLT} & {No match}\\
				\hline
				{APE}    &5.78    &\textbf{4.80}    &5.46    &\underline{5.42}    &6.63\\
				\hline
				{s0-10}  &\textbf{2.25}    &2.84    &3.65    &3.22    &\underline{2.47}\\
				\hline
				{s10-40} &6.26    &\textbf{6.08}    &6.52    &6.48    &\underline{6.18}\\
				\hline
				{s40+}   &19.03   &18.79   &\textbf{17.38}   &\underline{17.44}   &23.16\\
				\hline
			\end{tabular}}
		\end{center}
		\label{tab:Flow_Sintel}
	\end{table}
	
	\vspace{-0.2cm}

	\vspace{-0.2cm}
	\section{Conclusions}
	\label{sec:discussion}
	\vspace{-0.2cm}
	
	We have shown that the present work enables artificial neural networks to learn accurate matching from only ordinary videos. Though the performance evaluation indicates that MIND works surprisingly well in the expected unsupervised manner, it fails to outperform the existing empirically designed methods even in the resolution-reduced images. However, as stated in the very beginning, the aim of this work is to prove that it is possible to learn image matching without manual supervision, rather than to provide a more practicable algorithm for frame interpolation or image matching. Furthermore, we believe that the present unsupervised learning approach holds brilliant potential for the more natural solutions to similar low-level vision problems, such as optical flow, tracking and motion segmentation.
	
	Our future work focuses on making the present approach more applicable in real-world scenarios, in terms of both computational efficiency and reliability. It is also our hope that the present work helps to promote the concept of \textit{analysis by synthesis} towards broader acceptance.
	
\subsection*{Supplementary Material: \\Quantitative evaluation of frame interpolation}

We provide here the supplemental quantitative evaluation in terms of frame interpolation. The purposes are: 1. verifying that the trained CNNs performs quantitatively good frame interpolation; and 2. providing further evidence that the trained CNN holds good generalization ability.
	
	For the trained CNN (refereed to as MIND below), the interpolated images are simply the outputs of the forward propagation though the CNN. We compare the results to the traditional interpolation method using state-of-the-art optical flow, i.e. DeepFlow \cite{weinzaepfel2013deepflow} (initialized with DeepMatching). The interpolation algorithm used in the Middlebury benchmark \cite{baker2011database} is employed to synthesize the in-between images, given the optical flow fields obtained by DeepFlow. For simplicity, this approach is refereed to as DeepFlow below.

	The quantitative evaluations are performed on four image sequences: a representative sequence from KITTI RAW video\cite{Geiger2013IJRR}, one Sintel movie clip\footnote{Sintel, the Durian Open Movie Project. https://durian.blender.org/}, DICOM image sequence\footnote{The images are taken from a DICOM sample image set: http://www.osirix-viewer.com/datasets/. Alias Name: GRUSELAMBIX.} and RubberWhale sequence from the Middlebury optical flow benchmark \cite{baker2011database}. For each image sequence, MIND and DeepFlow are evaluated on each image triplet, where the first and third frames are taken as inputs and the second frame serves as ground truth interpolated frame.
	
        \subsection{Sample images}
	
\begin{figure}[t]
	\centering
	\includegraphics[width = 0.94\textwidth]{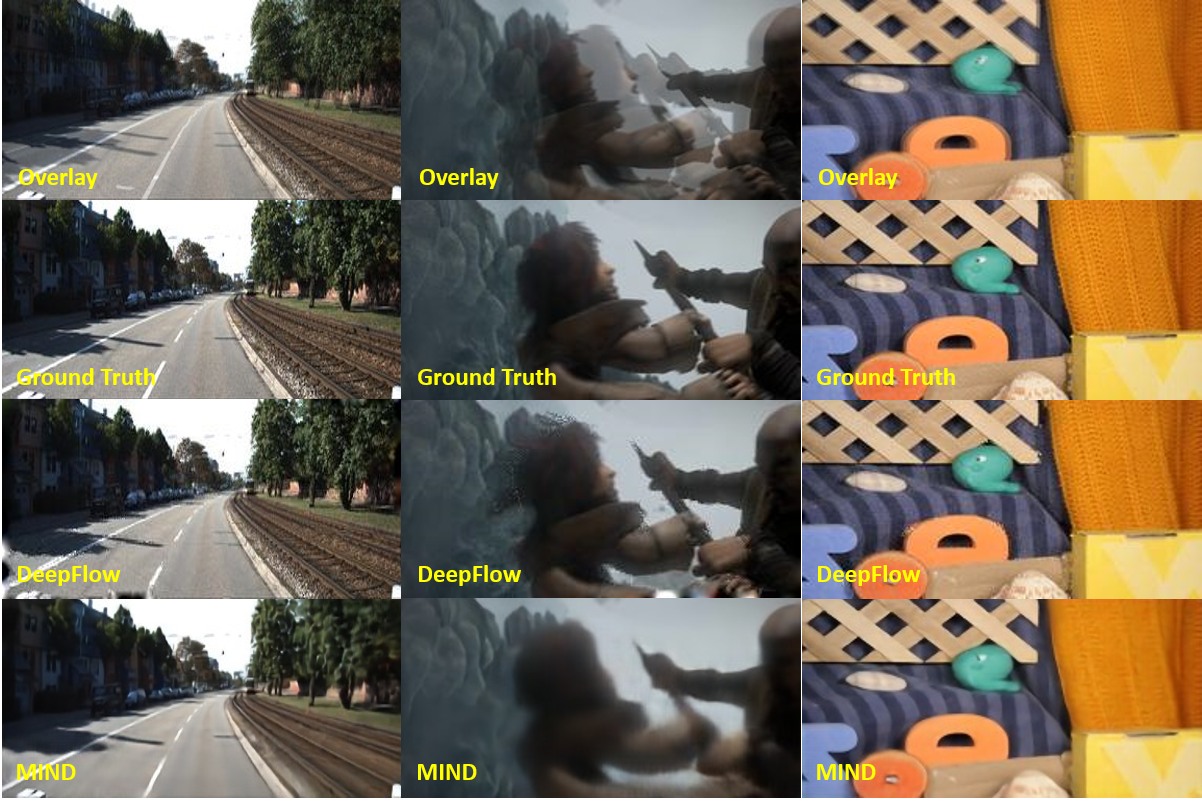}
	\caption{Examples of frame interpolation (best viewed in colour). From left to right: example on KITTI, Sintel and RubberWhale sequences, respectively. In each column, the first image is an overlay of the two input frames. The second one is the ground truth image. The third one and the fourth one is the interpolated image obtained by DeepFlow and MIND, respectively. For the first example, we use the CNN trained on KITTI itself. For all others, we use the CNN fine-tuned on Sintel data.}
	\label{fig:Interpolation_examples_SM}
	\end{figure}

\begin{figure}[!t]
		\centering
		\includegraphics[width = 0.94\textwidth]{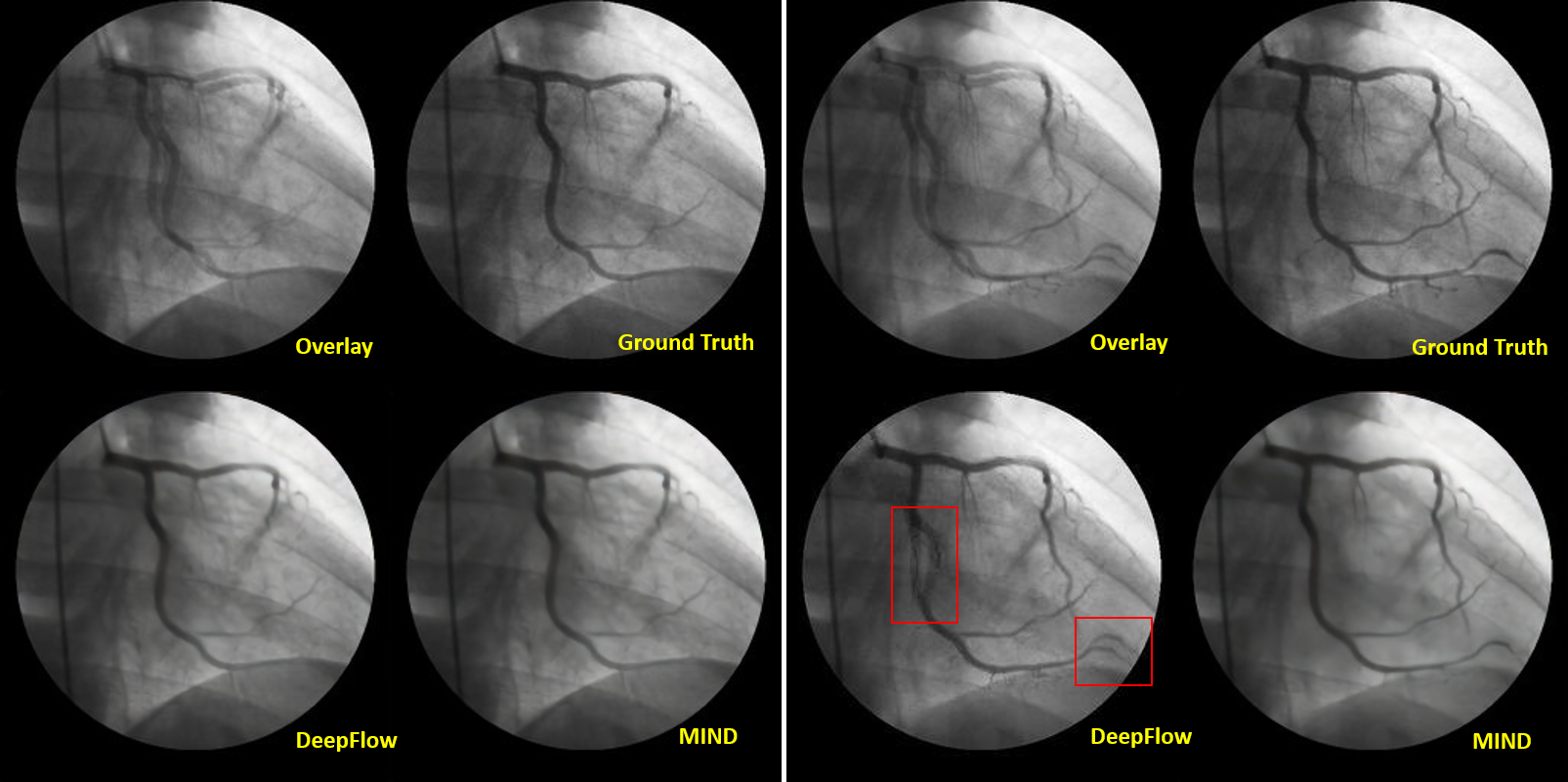}
		\caption{Two examples of frame interpolation on DICOM images. In each example, the first row shows the overlay of the two input frames and the ground truth frame. The second row shows the results obtained by DeepFlow and MIND. The inputs are scaled to the resolution of $256 \times 256 $. For MIND, we use the CNN fine-tuned on Sintel. Note that DeepFlow fails to cover the correct motion in the areas marked by the red squares.}
		\label{fig:Interpolation_dcm_SM}
	\end{figure}

	We first show some sample results for each image sequence. In Figure \ref{fig:Interpolation_examples_SM} and Figure \ref{fig:Interpolation_dcm_SM}, it can be seen that both MIND and DeepFlow work correctly for the task of frame interpolation. Please note that MIND works surprisingly well on DICOM and RubberWhale images, though it has never been trained with similar images. Notably, in the second example of DICOM images shown in Figure \ref{fig:Interpolation_dcm}, DeepFlow fails to cover the motion correctly, while MIND still obtains a good interpolated image.
	
	\vspace{+0.5cm}
	\subsection{Numerical results}
	Following \cite{baker2011database}, the interpolation error (IE) is defined as the root-mean-square (RMS) difference between the ground-truth image and the estimated interpolated image
	
	\begin{equation}
	IE = \left[\frac{1}{N}\sum_{\left(x, y\right)} \left(I(x, y)-I_{GT}(x, y)\right)^2\right]^{\frac{1}{2}}
	\end{equation}	
	where $N$ is the number of pixels. For color images, we take the L2 norm of the vector of RGB color differences.
	
	Furthermore, the normalized interpolation error (NE) between an interpolated image $I(x, y)$ and a ground truth image $I_{GT}(x, y)$ is defined as
	
	\begin{equation}
	NE = \left[\frac{1}{N}\sum_{\left(x, y\right)}\frac{\left(I(x, y)-I_{GT}(x, y)\right)^{2}}{\|\nabla I_{GT}(x, y)\|^{2}+\epsilon}\right]^{\frac{1}{2}}
        \end{equation}
	where $\epsilon=1.0$.

\begin{figure}[t]
	\centering
	\includegraphics[width = 0.95\textwidth]{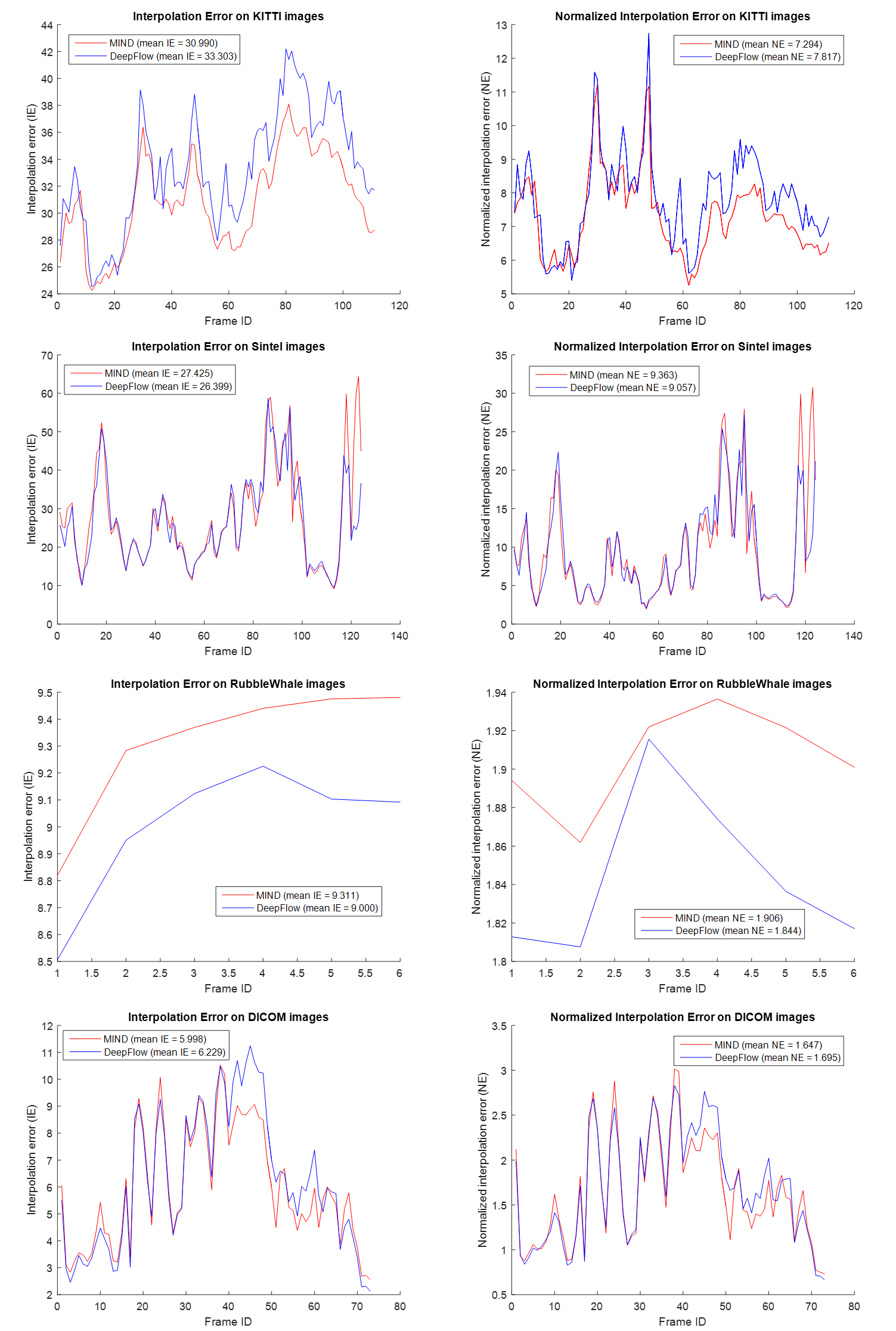}
	\caption{Interpolation error (IE) and Normalized interpolation error (NE) of interpolated images obtained by MIND and DeepFlow.}
	\label{fig:IE_NE}
	\end{figure}

	The numerical results of comparison between MIND and DeepFlow are reported in Table \ref{tab:interpolation_error} and Figure \ref{fig:IE_NE}. It can be seen that MIND works better on KITTI images than DeepFlow while failing to work well on Sintel images even though the CNN is fine-tuned using Sintel Movie clips. This fact is consistent with the evaluations on image matching and optical flow performance reported in our submission, which indicates that the current CNN cannot deal with Sintel images well mainly due to the existence of fast and complex motion.
	
	Regarding the generalization ability, it is further illustrated by the quantitative results that MIND learns indeed the ability to interpolate and match images, rather than only `remembering' the KITTI or Sintel-like images. MIND even does a better job than DeepFlow on DICOM images, which encourages us to explore its further applications. 

	\begin{table}
	\caption{Mean interpolation error (IE) and Mean normalized interpolation error (NE) of the interpolated images obtained by MIND and DeepFlow on each selected image sequence.}
        \begin{center}
	\scalebox{0.7}{
	\begin{tabular}{c|c  c  c  c}
	\hline
	& {\hspace{0.2cm}KITTI\hspace{0.2cm}} &{\hspace{0.2cm}Sintel\hspace{0.2cm}} & {\hspace{0.2cm}DICOM\hspace{0.2cm}} & {\hspace{0.2cm}RubbleWhale\hspace{0.2cm}}\\
	        \hline
                {IE (MIND)}     & 30.99    &27.43    &6.00    &9.31\\
                \hline
                {IE (DeepFlow)} & 33.30    &26.40    &6.23    &9.00\\
                \hline
                {NE (MIND)}     & 7.29    &9.36      &1.65    &1.91\\
                \hline
                {NE (DeepFlow)} & 7.82    &9.06      &1.70    &1.84\\
                \hline
                \end{tabular}}
		\end{center}
	\label{tab:interpolation_error}
	\end{table}
		
\clearpage

\bibliographystyle{splncs}
\bibliography{myBib}
\end{document}